\documentclass[10pt,twocolumn,letterpaper]{article}

\usepackage{iccv}
\usepackage{times}
\usepackage{epsfig}
\usepackage{graphicx}
\usepackage{amsmath}
\usepackage{amssymb}
\usepackage{color}
\usepackage{siunitx}
\usepackage[utf8]{inputenc}
\usepackage[inline]{enumitem}
\usepackage[nolist,nohyperlinks]{acronym}
\usepackage[breaklinks=true,bookmarks=false]{hyperref}

\def\iccvPaperID{1750} 

\iccvfinalcopy

\ificcvfinal\fi
\author{Christian Bartz \qquad Haojin Yang \qquad Christoph Meinel\\
Hasso Plattner Institute, University of Potsdam\\
Prof.-Dr.-Helmert Straße 2-3\\ 14482 Potsdam, Germany\\
{\tt\small \{christian.bartz, haojin.yang, meinel\}@hpi.de}
}

\begin{document}
	\title{STN-OCR: A single Neural Network for Text Detection and Text Recognition}
	\maketitle

	\begin{acronym}
		\acro{OCR}{Optical Character Recognition}
		\acro{CNN}{Convolutional Neural Network}
		\acro{RNN}{Recurrent Neural Network}
		\acro{DNN}{Deep Neural Network}
		\acro{BLSTM}{Bidirectional Long-Short Term Memory}
		\acro{CTC}{Connectionist Temporal Classification}
		\acro{FSNS}{French Street Name Signs}
		\acro{SGD}{Stochastic Gradient Descent}
	\end{acronym}

	\begin{abstract}
		Detecting and recognizing text in natural scene images is a challenging, yet not completely solved task.
		In recent years several new systems that try to solve at least one of the two sub-tasks (text detection and text recognition) have been proposed.
		In this paper we present STN-OCR, a step towards semi-supervised neural networks for scene text recognition that can be optimized end-to-end.
		In contrast to most existing works that consist of multiple deep neural networks and several pre-processing steps we propose to use a single deep neural network that learns to detect and recognize text from natural images in a semi-supervised way.
		STN-OCR is a network that integrates and jointly learns a spatial transformer network \cite{Jaderberg2015Spatial}, that can learn to detect text regions in an image, and a text recognition network that takes the identified text regions and recognizes their textual content.
		We investigate how our model behaves on a range of different tasks (detection and recognition of characters, and lines of text).
		Experimental results on public benchmark datasets show the ability of our model to handle a variety of different tasks, without substantial changes in its overall network structure.
	\end{abstract}

	\section{Introduction}

Text is ubiquitous in our daily lifes.
Text can be found on documents, road signs, billboards, and other objects like cars or telephones.
Automatically detecting and reading text from natural scene images is an important part of systems that can be used for several challenging tasks such as image-based machine translation, autonomous cars or image/video indexing.
In recent years the task of detecting text and recognizing text in natural scenes has seen much interest from the computer vision and document analysis community.
Furthermore recent breakthroughs \cite{He2016Deep,Jaderberg2015Spatial,Redmon2016You,Ren2015Faster} in other areas of computer vision enabled the creation of even better scene text detection and recognition systems than before \cite{Bigorda2016Textproposalsa,Gupta2016Syntheticb,Shi2016Robust}.
Although the problem of \ac{OCR} can be seen as solved for printed document texts it is still challenging to detect and recognize text in natural scene images.
Images containing natural scenes exhibit large variations of illumination, perspective distortions, image qualities, text fonts, diverse backgrounds, \etc.

The majority of existing research works developed end-to-end scene text recognition systems that consist of complex two-step pipelines, where the first step is to detect regions of text in an image and the second step is to recognize the textual content of that identified region.
Most of the existing works only concentrate on one of these two steps.

\begin{figure}[t]
	\begin{center}
		\includegraphics[width=1.0\linewidth]{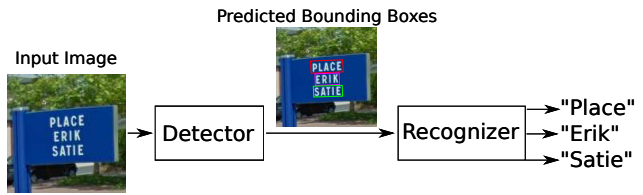}
	\end{center}
	   \caption{Schematic overview of our proposed system. The input image is fed to a single neural network that consists of a text detection part and a text recognition part. The text detection part learns to detect text in a semi-supervised way, by being jointly trained with the recognition part.}
	\label{fig:schematic_of_system}
\end{figure}

In this paper, we present a solution that consists of a single \ac{DNN} that can learn to detect and recognize text in a semi-supervised way.
This is contrary to existing works, where text detection and text recognition systems are trained separately in a fully-supervised way.
Recent work \cite{Dai2016InstanceAware} showed that \acp{CNN} are capable of learning how to solve complex multi-task problems, while being trained in an end-to-end manner.
Our motivation is to use these capabilities of \acp{CNN} and create an end-to-end scene text recognition system that behaves more like a human by dividing the task at hand into smaller subtasks and solving these subtask independently from each other.
In order to achieve this behavior we learn a single \ac{DNN} that is able to divide the input image into subtasks (single characters, words or even lines of text) and solve these subtasks independently of each other.
This is achieved by jointly learning a localization network that uses a recurrent spatial transformer \cite{Jaderberg2015Spatial,Snderby2015Recurrent} as attention mechanism and a text recognition network (see \autoref{fig:schematic_of_system} for a schematic overview of the system).
In this setting the network only receives the image and the labels for the text contained in that image as input.
The localization of the text is learned by the network itself, making this approach semi-supervised.

Our contributions are as follows:
\begin{enumerate*}[label={(\arabic*)}]
	\item We present a system that is a step towards solving end-to-end scene text recognition by integrating spatial transformer networks.
	\item We train our proposed system end-to-end in a semi-supervised way.
	\item We demonstrate that our approach is able to reach state-of-the-art/competitive performance on a range of standard scene text detection and recognition benchmarks.
	\item We provide our code\footnote{\url{https://github.com/Bartzi/stn-ocr}} and trained models\footnote{\url{https://bartzi.de/research/stn-ocr}} to the research community.
\end{enumerate*}

This paper is structured in the following way:
In \autoref{sec:related_work} we outline work of other researchers related to ours.
Section \ref{sec:proposed_system} describes our proposed system in detail and provides best practices on how to train such a system.
We show and discuss our results on standard benchmark datasets in \autoref{sec:experiments} and conclude our findings in \autoref{sec:conclusion}.

	\section{Related Work}
\label{sec:related_work}

Over the course of years a rich environment of different approaches to scene text detection and recognition have been developed and published.
Nearly all systems use a two-step process for performing end-to-end recognition of scene text.
The first step is to detect regions of text and extract these regions from the input image.
The second step is to recognize the textual content and return the text strings of these extracted text regions.

It is further possible to divide these approaches into three broad categories:
\begin{enumerate*}[label={(\arabic*)}]
	\item Systems relying on hand crafted features and human knowledge for text detection and text recognition.
	\item Systems using deep learning approaches together with hand crafted features, or two different deep networks for each of the two steps.
	\item Systems that do not consist of a two step approach but rather perform text detection and recognition using a single deep neural network.
\end{enumerate*}
We will discuss some of these systems for each category below.

\paragraph{Hand Crafted Features}
	In the beginning methods based on hand crafted features and human knowledge have been used to perform text detection.
	These systems used features like MSERs \cite{Neumann2010Method}, Stroke Width Transforms \cite{Epshtein2010Detecting} or HOG-Features \cite{Wang2011EndToEnd} to identify regions of text and provide them to the text recognition stage of the system.
	In the text recognition stage sliding window classifiers \cite{Mishra2012Scene} and ensembles of SVMs \cite{Yao2014Strokelets} or k-Nearest Neighbor classifiers using HOG features \cite{Wang2010Word} were used.
	All of these approaches use hand crafted features that have a large variety of hyper parameters that need expert knowledge to correctly tune them for achieving the best results.

\paragraph{Deep Learning Approaches}
	More recent systems exchange approaches based on hand crafted features in one or both steps of end-to-end recognition systems by approaches using \acp{DNN}.
	Gómez and Karatzas \cite{Bigorda2016Textproposalsa} propose a text-specific selective search algorithm that, together with a \ac{DNN}, can be used to detect (distorted) text regions in natural scene images.
	Gupta \etal \cite{Gupta2016Syntheticb} propose a text detection model based on the YOLO-Architecture \cite{Redmon2016You} that uses a fully convolutional deep neural network to identify text regions.
	The text regions identified by these approaches can then be used as input for further systems based on \acp{DNN} that perform text recognition.

	Bissacco \etal \cite{Bissacco2013Photoocr} propose a complete end-to-end architecture that performs text detection using hand crafted features. The identified text regions are binarized and then used as input to a deep fully connected neural network that classifies each found character independently.
	Jaderberg \etal \cite{Jaderberg2015Reading,Jaderberg2014Deep} propose several systems that use deep neural networks for text detection and text recognition.
	In \cite{Jaderberg2014Deep} Jaderberg \etal propose a sliding window text detection approach that slides a convolutional text detection model across the image in multiple resolutions.
	The text recognition stage uses a single character \ac{CNN}, which is slided across the identified text region.
	This \ac{CNN} shares its weights with the \ac{CNN} used for text detection.
	In \cite{Jaderberg2015Reading} Jaderberg \etal propose to use a region proposal network with an extra bounding box regression CNN for text detection and a CNN that takes the whole text region as input and performs classification across a pre-defined dictionary of words, making this approach only applicable to one given language.

	Goodfellow \etal \cite{Goodfellow2014MultiDigit} propose a text recognition system for house numbers, that has been refined by Jaderberg \etal \cite{Jaderberg2014Deepa} for unconstrained text recognition.
	This system uses a single \ac{CNN}, which takes the complete extracted text region as input, and provides the text contained in that text region.
	This is achieved by having one independent classifier for each possible character in the given word.
	Based on this idea He \etal \cite{He2016Reading} and Shi \etal \cite{Shi2016EndToEnd,Shi2016Robust} propose text recognition systems that treat the recognition of characters from the extracted text region as a sequence recognition problem.
	He \etal \cite{He2016Reading} use a naive sliding window approach that creates slices of the text region, which are used as input to their text recognition \ac{CNN}.
	The features produced by the text recognition \ac{CNN} are used as input to a \ac{RNN} that predicts the sequence of characters.
	In our experiments on pure scene text recognition (see section \ref{ssec:icdar2013_experiments} for more information) we use a similar approach, but our system uses a more sophisticated sliding window approach, where the choice of the sliding windows is automatically learned by the network and not engineered by hand.
	Shi \etal \cite{Shi2016EndToEnd} utilize a CNN that uses the complete text region as input and produces a sequence of feature vectors, which are fed to a RNN that predicts the sequence of characters in the extracted text region.
	This approach generates a fixed number of feature vectors based on the width of the text region.
	That means for a text region that only contains a few characters, but has the same width as a text region with sufficently more characters, this approach will produce the same amount of feature vectors used as input to the RNN. In our pure text recognition experiments we utilized the strength of our approach to learn to attend to the most important information in the extracted text region, hence producing only as many feature vectors as necessary.
	Shi \etal \cite{Shi2016Robust} improve their approach by firstly adding an extra step that utilizes the rectification capabilities of Spatial Transformer Networks \cite{Jaderberg2015Spatial} for rectifying the extracted text line.
	Secondly they added a soft-attention mechanism to their network that helps to produce the sequence of characters in the input image.
	In their work Shi \etal make use of Spatial Transformers as an extra pre-processing step to make it easier for the recognition network to recognize the text in the image. In our system we use the Spatial Transformer as a core building block for detecting text in a semi-supervised way.

\paragraph{End-to-End trainable Approaches}
	The presented systems always use a two-step approach for detecting and recognizing text from scene text images.
	Although recent approaches make use of deep neural networks they are still using a huge amount of hand crafted knowledge in either of the steps or at the point where the results of both steps are fused together.
	Smith \etal \cite{Smith2016EndToEnd} propose an end-to-end trainable system that is able to detect and recognize text on french street name signs, using a single \ac{DNN}.
	In contrast to our system it is not possible for the system to provide the location of the text in the image, only the textual content can be extracted.
	Furthermore the attention mechanism used in our approach shows a more human-like behaviour because is sequentially localizes and recognizes text from the given image.

	\section{Proposed System}
\label{sec:proposed_system}

A human trying to find and read text will do so in a sequential manner. The first action is to put attention on a line of text, read each character sequentially and then attend to the next line of text.
Most current end-to-end systems for scene text recognition do not behave in that way.
These systems rather try to solve the problem by extracting all information from the image at once.
Our system first tries to attend sequentially to different text regions in the image and then recognize the textual content of each text region.
In order to this we created a simple \ac{DNN} consisting of two stages:
\begin{enumerate*}[label={(\arabic*)}]
	\item text detection
	\item text recognition
\end{enumerate*}.
In this section we will introduce the attention concept used by the text detection stage and the overall structure of the proposed system.
We also report best practices for successfully training such a system.

\subsection{Detecting Text with Spatial Transformers}
\label{subsec:ps_spatial_transformer_networks}

A spatial transformer proposed by Jaderberg \etal \cite{Jaderberg2015Spatial} is a differentiable module for \acp{DNN} that takes an input feature map $I$ and applies a spatial transformation to this feature map, producing an output feature map $O$.
Such a spatial transformer module is a combination of three parts.
The first part is a localisation network computing a function $f_{loc}$, that predicts the parameters $\theta$ of the spatial transformation to be applied.
These predicted parameters are used in the second part to create a sampling grid that defines which features of the input feature map should be mapped to the output feature map.
The third part is a differentiable interpolation method that takes the generated sampling grid and produces the spatially transformed output feature map $O$.
We will shortly describe each component in the following paragraphs.

\begin{figure}[t]
	\begin{center}
		\includegraphics[width=0.9\linewidth]{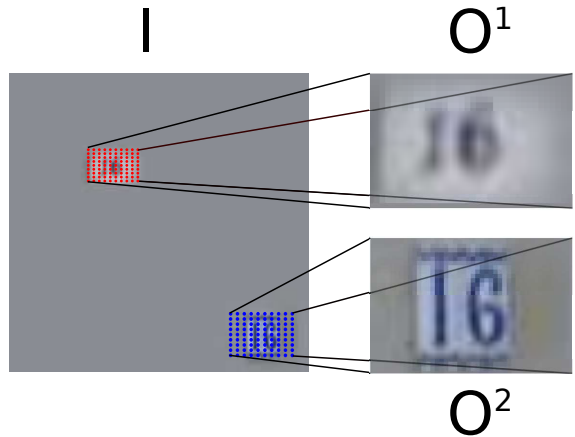}
	\end{center}
	\caption{Operation method of grid generator and image sampler. First the grid generator uses the $N$ affine transformation matrices $A^{n}_{\theta}$ to create $N$ equally spaced sampling grids (red and blue grids on the left side). These sampling grids are used by the image sampler to extract the image pixels at that location, in this case producing the two output images $O^1$ and $O^2$. The corners of the generated sampling grids provide the vertices of the bounding box for each text region that has been found by the network.}
	\label{fig:transformation_params_overview}
\end{figure}

\paragraph{Localization Network}
	The localization network takes the input feature map $I \in \mathbb{R}^{C \times H \times W}$, with $C$ channels, height $H$ and width $W$ and outputs the parameters $\theta$ of the transformation that shall be applied.
	In our system we use the localization network ($f_{loc}$) to predict $N$ two-dimensional affine transformation matrices $A^{n}_{\theta}$, where $n \in \{0, \ldots, N - 1\}$:
	\begin{equation}
		f_{loc}(I) = A^{n}_{\theta} = 
		\begin{bmatrix}
			\theta^{n}_1 & \theta^{n}_2 & \theta^{n}_3 \\
			\theta^{n}_4 & \theta^{n}_5 & \theta^{n}_6 \\
		\end{bmatrix}
	\end{equation}

	$N$ is thereby the number of characters, words or textlines the localization network shall localize.
	The affine transformation matrices predicted in that way allow the network to apply translation, rotation, zoom and skew to the input image, hence the network learns to produce transformation parameters that can zoom on characters, words or text lines that are to be extracted from the image.

	In our system the $N$ transformation matrices $A^{n}_{\theta}$ are produced by using a feed-forward \ac{CNN} together with a \ac{RNN}. Each of the $N$ transformation matrices is computed using the hidden state $h_n$ for each time-step of the \ac{RNN}:
	\begin{align}
		c &= f^{conv}_{loc}(I) \\
		h_n &= f^{rnn}_{loc}(c, h_{n-1}) \\
		A^{n}_{\theta} &= g_{loc}(h_n)
	\end{align}
	where $g_{loc}$ is another feed-forward network, and each transformation matrix $A^{n}_{\theta}$ is conditioned on the globally extracted convolutional features ($f^{conv}_{loc}$) together with the hidden state of the previously performed time-step.

\begin{figure*}[t]
	\begin{center}
		\includegraphics[width=0.9\textwidth]{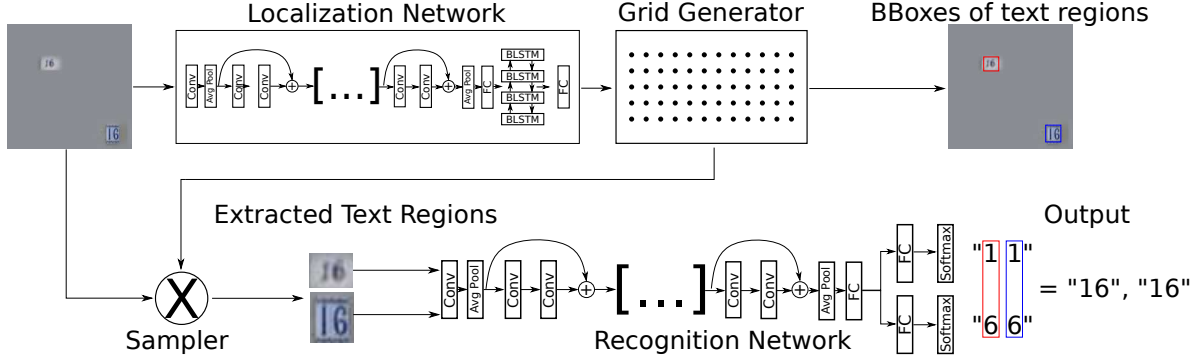}
	\end{center}
	\caption{The network used in our work consists of two major parts. The first is the localization network that takes the input image and predicts $N$ transformation matrices, that are applied to $N$ identical grids, forming $N$ different sampling grids. The generated sampling grids are used in two ways:
	(1) for calculating the bounding boxes of the identified text regions
	(2) for sampling the input image with $N$ sampling grids to extract $N$ text regions. 
	The $N$ extracted text images are then used in the recognition network to perform text recognition.
	The whole system is trained end-to-end by only supplying information about the text labels for each text region.}
	\label{fig:localization_net_structure}
\end{figure*}

	The \ac{CNN} in the localization network used by us is a variant of the well known ResNet by He \etal \cite{He2016Deep}.
	We use a variant of ResNet because we found that with this network structure our system learns faster and more successful, as compared to experiments with other network structures like the VGGNet \cite{Simonyan2015Very}.
	We argue that this is due to the fact that the residual connections of the ResNet help with retaining a strong gradient down to the very first convolutional layers.
	In addition to the structure we also used Batch Normalization \cite{Ioffe2015Batcha} for all our experiments.
	The \ac{RNN} used in the localization network is a \ac{BLSTM} \cite{Graves2013Hybrid,Hochreiter1997Long} unit.
	This \ac{BLSTM} is used to generate the hidden states $h_n$, which in turn are used to predict the affine transformation matrices.
	We used the same structure of the network for all our experiments we report in \autoref{sec:experiments}.
	\autoref{fig:localization_net_structure} provides a structural overview of this network.

\paragraph{Grid Generator}
	The grid generator uses a regularly spaced grid $G_o$ with coordinates $y_{h_o},x_{w_o}$, of height $H_o$ and width $W_o$, together with the affine transformation matrices $A^{n}_{\theta}$ to produce $N$ regular grids $G^n_i$ of coordinates $u^n_{i},v^n_{j}$ of the input feature map $I$, where $i \in H_o$ and $j \in W_o$:
	\begin{equation}
		\begin{pmatrix}
			u^n_{i} \\
			v^n_{j}
		\end{pmatrix}
		= A^{n}_{\theta}
		\begin{pmatrix}
			x_{w_o} \\
			y_{h_o} \\
			1
		\end{pmatrix}
		= \begin{bmatrix}
			\theta^{n}_1 & \theta^{n}_2 & \theta^{n}_3 \\
			\theta^{n}_4 & \theta^{n}_5 & \theta^{n}_6 \\
		\end{bmatrix}
		\begin{pmatrix}
			x_{w_o} \\
			y_{h_o} \\
			1
		\end{pmatrix}
	\end{equation}
	During inference we can extract the $N$ resulting grids $G^n_i$ which contain the bounding boxes of the text regions found by the localization network.
	Height $H_o$ and width $W_o$ can be chosen freely and if they are lower than height $H$ or width $W$ of the input feature map $I$ the grid generator is producing a grid that performs a downsampling operation in the next step.
	
\paragraph{Image Sampling}
	The $N$ sampling grids $G^n_i$ produced by the grid generator are now used to sample values of the feature map $I$ at their corresponding coordinates $u^n_{i},v^n_{j}$ for each $n \in N$. Naturally these points will not always perfectly align with the discrete grid of values in the input feature map.
	Because of that we use bilinear sampling that extracts the value at a given coordinate by bilinear interpolating the values of the nearest neighbors. With that we define the values of the $N$ output feature maps $O^n$ at a given location $i,j$ where $i \in H_o$ and $j \in W_o$:
	\begin{equation}
		O^n_{ij} = \sum^{H_o}_h \sum^{W_o}_w I_{hw} max(0, 1 - \lvert u^n_{i} - h \rvert) max(0, 1 - \rvert v^n_{j} - w \rvert)
	\end{equation}
	This bilinear sampling is (sub-)differentiable, hence it is possible to propagate error gradients to the localization network by using standard backpropagation.

The combination of localization network, grid generator and image sampler forms a spatial transformer and can in general be used in every part of a \ac{DNN}.
In our system we use the spatial transformer as the first step of our network. The localization network receives the input image as input feature map and produces a set of affine transformation matrices that are used by the grid generator to calculate the position of the pixels that shall be sampled by the bilinear sampling operation.\\

\subsection{Text Recognition Stage}

The image sampler of the text detection stage produces a set of $N$ regions that are extracted from the original input image.
The text recognition stage (a structural overview of this stage can be found in \autoref{fig:localization_net_structure}) uses each of these $N$ different regions and processes them independently of each other.
The processing of the $N$ different regions is handled by a \ac{CNN}.
This \ac{CNN} is also based on the ResNet architecture as we found that we could only achieve good results if we use a variant of the ResNet architecture for our recognition network.
We argue that using a ResNet in the recognition stage is even more important than in the detection stage, because the detection stage needs to receive strong gradients from the recognition stage in order to successfully update the weights of the localization network.
The \ac{CNN} of the recognition stage predicts a probability distribution $\hat{y}$ over the label space $L_{\epsilon}$, where $L_{\epsilon} = L \cup \{\epsilon\}$, with $L = \{0-9a-z\}$ and $\epsilon$ representing the blank label.
Depending on the task this probability distribution is either generated by a fixed number of $T$ softmax classifiers, where each softmax classifier is used to predict one character of the given word:
\begin{align}
	x^n &= O^n \\
	\hat{y}^n_t &= softmax(f_{rec}(x^n)) \\
	\hat{y}^n &= \sum_{t=1}^
	T \hat{y}^n_t
\end{align}
where $f_{rec}(x)$ is the result of applying the convolutional feature extractor on the sampled input $x$.

Another possibility is to train the network using \ac{CTC} \cite{Graves2006Connectionist} and retrieve the most probable labeling by setting $\hat{y}$ to be the most probable labeling path $\pi$, that is given by:
\begin{align}
	p(\pi|x^n) &= \prod^{T}_{t=1} \hat{y}^n_{\pi_t}, \forall \pi \in L^T_{\epsilon} \\
	\hat{y}^n_t &= \text{argmax} p(\pi|x^n) \\
	\hat{y}^n &= B(\sum^T_{t=1} \hat{y}^n_t)
\end{align}
with $L^T_{\epsilon}$ being the set of all labels that have the length $T$ and $p(\pi|x^n)$ being the probability that path $\pi \in L^T_{\epsilon}$ is predicted by the \ac{DNN}. $B$ is a function that removes all predicted blank labels and all repeated labels (e.g. $B(\text{-IC-CC-V}) = B(\text{II--CCC-C--V-}) = \text{ICCV}$).

\subsection{Model Training}

The training set $X$ used for training the model consists of a set of input images $I$ and a set of text labels $L_I$ for each input image.
We do not use any labels for training the text detection stage.
This stage is learning to detect regions of text only by using the error gradients obtained by either calculating the cross-entropy loss or the \ac{CTC} loss of the predictions and the textual labels.
During our experiments we found that, when trained from scratch, a network that shall detect and recognize more than two text lines does not converge.
The solution to this problem is to perform a series of pre-training steps where the difficulty is gradually increasing.
Furthermore we find that the optimization algorithm chosen to train the network has a great influence on the convergence of the network.
We found that it is beneficial to use \ac{SGD} for pre-training the network on a simpler task and Adam \cite{Kingma2015Adam} for finetuning the already pre-trained network on images with more text lines.
We argue that \ac{SGD} performs better during pre-training because the learning rate $\eta$ is kept constant during a longer period of time, which enables the text detection stage to explore the input images and better find text regions.
With decreasing learning rate the updates in the detection stage become smaller and the text detection stage (ideally) settles on already found text regions.
At the same time the text recognition network can start to use the extracted text regions and learn to recognize the text in that regions.
While training the network with \ac{SGD} it is important to note that choosing a too high learning rate will result in divergence of the model early on.
We found that using initial learning rates between $1^{-5}$ and $5^{-7}$ tend to work in nearly all cases, except in cases where the network should only be fine-tuned.
Here we found that using Adam is the more reliable choice, as Adam chooses the learning rate for each parameter in an adaptive way and hence does not allow the detection network to explore as radically as it does when using \ac{SGD}.

	\section{Experiments}
\label{sec:experiments}

In this section we evaluate our presented network architecture on several standard scene text detection/recognition datasets.
We present the results of experiments for three different datasets, where the difficulty of the task at hand increases for each dataset.
We first begin with experiments on the SVHN dataset \cite{Netzer2011Reading}, that we used to prove that our concept as such is feasible.
The second type of dataset we performed experiments on were datasets for focused scene text recognition, where we explored the performance of our model, when it comes to find and recognize single characters.
The third dataset we exerimented with was the \acf{FSNS} dataset \cite{Smith2016EndToEnd}, which is the most challenging dataset we used, as this dataset contains a vast amount of irregular, low resolution text lines that are more difficult to locate and recognize than text lines from the SVHN dataset.
We begin this section by introducing our experimental setup.
We will then present the results and characteristics of the experiments for each of the aforementioned datasets.

\subsection{Experimental Setup}
\label{ssec:experimental_setup}

\paragraph{Localization Network}
The localization network used in every experiment is based on the ResNet architecture \cite{He2016Deep}.
The input to the network is the image where text shall be localized and later recognized.
Before the first residual block the network performs a $3 \times 3$ convolution followed by a $2 \times 2$ average pooling layer with stride 2.
After these layers three residual blocks with two $3 \times 3$ convolutions, each followed by batch normalization \cite{Ioffe2015Batcha}, are used.
The number of convolutional filters is 32, 48 and 48 respectively and ReLU \cite{Nair2010Rectified} is used as activation function for each convolutional layer.
A $2 \times 2$ max-pooling with stride 2 follows after the second residual block.
The last residual block is followed by a $5 \times 5$ average pooling layer and this layer is followed by a \ac{BLSTM} with 256 hidden units.
For each time step of the \ac{BLSTM} a fully connected layer with 6 hidden units follows.
This layer predicts the affine transformation matrix, that is used to generate the sampling grid for the bilinear interpolation.
As rectification of scene text is beyond the scope of this work we disabled skew and rotation in the affine transformation matrices by setting the according parameters to 0.
We will discuss the rectification capabilities of Spatial Transformers for scene text detection in our future work.

\paragraph{Recognition Network}
The inputs to the recognition network are $N$ crops from the original input image that represent the text regions found by the localization network.
The recognition network has the same structure as the localization network, but the number of convolutional filters is higher.
The number of convolutional filters is 32, 64 and 128 respectively.
Depending on the experiment we either used an ensemble of $T$ independent softmax classifiers as used in \cite{Goodfellow2014MultiDigit} and \cite{Jaderberg2014Deep}, where $T$ is the maximum length that a word may have, or we used \ac{CTC} with best path decoding as used in \cite{He2016Reading} and \cite{Shi2016EndToEnd}.

\paragraph{Implementation}
We implemented all our experiments using MXNet \cite{Chen2015Mxnet}. We conduted all our experiments on a work station which has an Intel(R) Core(TM) i7-6900K CPU, 64 GB RAM and 4 TITAN X (Pascal) GPUs.

\subsection{Experiments on the SVHN dataset}
\label{ssec:svhn_experiments}

With our first experiments on the SVHN dataset \cite{Netzer2011Reading} we wanted to prove that our concept works and can be used with real world data.
We therefore first conducted experiments similar to the experiments in \cite{Jaderberg2015Spatial} on SVHN image crops with a single house number in each image crop, that is centered around the number and also contains background noise.
\autoref{tab:svhn_results} shows that we are able to reach competitive recognition accuracies.

\begin{table}
	\begin{center}
		\begin{tabular}{|l|c|}
			\hline
			Method & 64px \\
			\hline
			Maxout CNN \cite{Goodfellow2014MultiDigit} & 96 \\
			ST-CNN \cite{Jaderberg2015Spatial} & 96.3 \\
			\hline
			Ours & 95.2 \\
			\hline
		\end{tabular}
	\end{center}
	\caption{Sequence recognition accuracies on the SVHN dataset. When recognizing house number on crops of $64 \times 64$ pixels, following the experimental setup of \cite{Goodfellow2014MultiDigit}}
	\label{tab:svhn_results}
\end{table}

\begin{figure}[t]
	\begin{center}
		\includegraphics[width=0.9\linewidth]{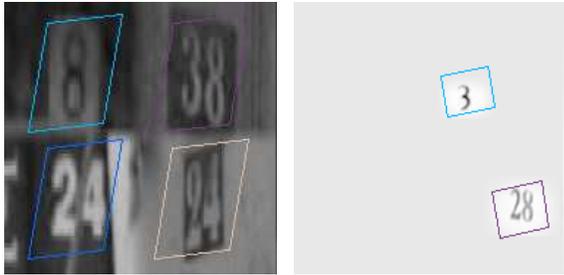}
	\end{center}
	\caption{Samples from our generated datasets, including BBoxes predicted by our model. \textit{Left:} Sample from regular grid dataset, \textit{Right:} Sample from dataset with randomly positioned house numbers.}
	\label{fig:svhn_grid_dataset}
\end{figure}

Based on this experiment we wanted to determine whether our model is able to detect different lines of text that are arranged in a regular grid or placed at random locations in the image.
In \autoref{fig:svhn_grid_dataset} we show samples from two purpose build datasets\footnote{datasets are available here: \url{https://bartzi.de/research/stn-ocr}} that we used for our other experiments based on SVHN data.
We found that our network performs well on the task of finding and recognizing house numbers that are arranged in a regular grid.
An interesting observation we made during training on this data was that we were able to achieve our best results when we did two training steps.
The first step was to train the complete model from scratch (all weights initialized randomly) and then train the model with the same data again, but this time with the localization network pre-initialized with the weights obtained from the last training and the recognition net initialized with random weights.
This strategy leads to better localization results of the localization network and hence improved recognition results.

During our experiments on the second dataset, created by us, we found that it is not possible to train a model from scratch, that can find and recognize more than two textlines that are scattered across the whole image.
It is possible to train such a network by first training the model on easier tasks first (few textlines, textlines closer to the center of the image) and then increase the difficulty of the task gradually.
In the supplementary material we provide short video clips that show how the network is exploring the image while learning to detect text for a range of different experiments.

\subsection{Experiments on Robust Reading Datasets}
\label{ssec:icdar2013_experiments}

\begin{figure}[t]
	\begin{center}
		\includegraphics[width=0.9\linewidth]{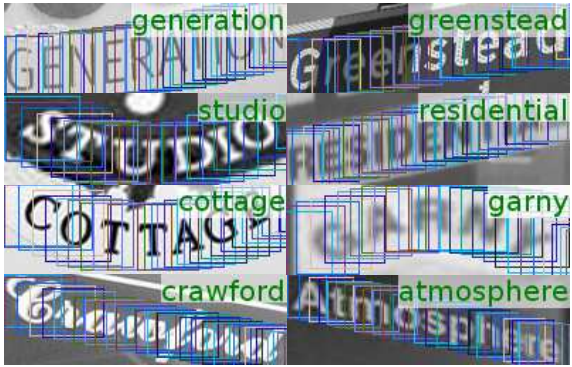}
	\end{center}
	\caption{Samples from ICDAR, SVT and IIIT5K datasets that show how well our model finds text regions and is able to follow the slope of the words.}
	\label{fig:robust_reading_text_localization}
\end{figure}

In our next experiments we used datasets where text regions are aleady cropped from the input images.
We wanted to see whether our text localization network can be used as an intelligent sliding window generator that adopts to irregularities of the text in the cropped text region.
Therefore we trained our recognition model using \ac{CTC} on a dataset of synthetic cropped word images, that we generated using our own data generator, that works similar to the data generator introduced by Jaderberg \etal \cite{Jaderberg2014Synthetic}.
In \autoref{tab:focused_scene_text_results} we report the recognition results of our model on the ICDAR 2013 robust reading \cite{Karatzas2013Icdar}, the Street View Text (SVT) \cite{Wang2011EndToEnd} and the IIIT5K \cite{Mishra2012Scene} benchmark datasets.
For evaluation on the ICDAR 2013 and SVT datasets, we filtered all images that contain non-alphanumeric characters and discarded all images that have less than 3 characters as done in \cite{Shi2016Robust,Wang2011EndToEnd}.
We obtained our final results by post-processing the predictions using the standard hunspell english (en-US) dictionary.
Overall we find that our model achieves state-of-the-art performance for unconstrained recognition models on the ICDAR 2013 and IIIT5K dataset and competitive performance on the SVT dataset.
In \autoref{fig:robust_reading_text_localization} we show that our model learns to follow the slope of the individual text regions, proving that our model produces sliding windows in an intelligent way.

\begin{table}
	\begin{center}
		\begin{tabular}{|l|c|c|c|}
			\hline
			Method & ICDAR 2013 & SVT & IIIT5K \\
			\hline
			Photo-OCR \cite{Bissacco2013Photoocr} & 87.6 & 78.0 & -\\
			CharNet \cite{Jaderberg2014Deepa} & 81.8 & 71.7 & -\\
			DictNet* \cite{Jaderberg2015Reading} & \textbf{90.8} & 80.7 & - \\
			CRNN \cite{Shi2016EndToEnd} & 86.7 & 80.8 & 78.2 \\
			RARE \cite{Shi2016Robust} & 87.5 & \textbf{81.9} & 81.9 \\
			\hline
			Ours & \textbf{90.3} & 79.8 & \textbf{86} \\
			\hline
		\end{tabular}
	\end{center}
	\caption{Recognition accuracies on the ICDAR 2013, SVT and IIIT5K robust reading benchmarks. Here we only report results that do not use per image lexicons. (*\cite{Jaderberg2015Reading} is not lexicon-free in the strict sense as the outputs of the network itself are constrained to a 90k dictionary.)}
	\label{tab:focused_scene_text_results}
\end{table}

\begin{figure*}[t]
	\begin{center}
		\includegraphics[width=0.99\linewidth]{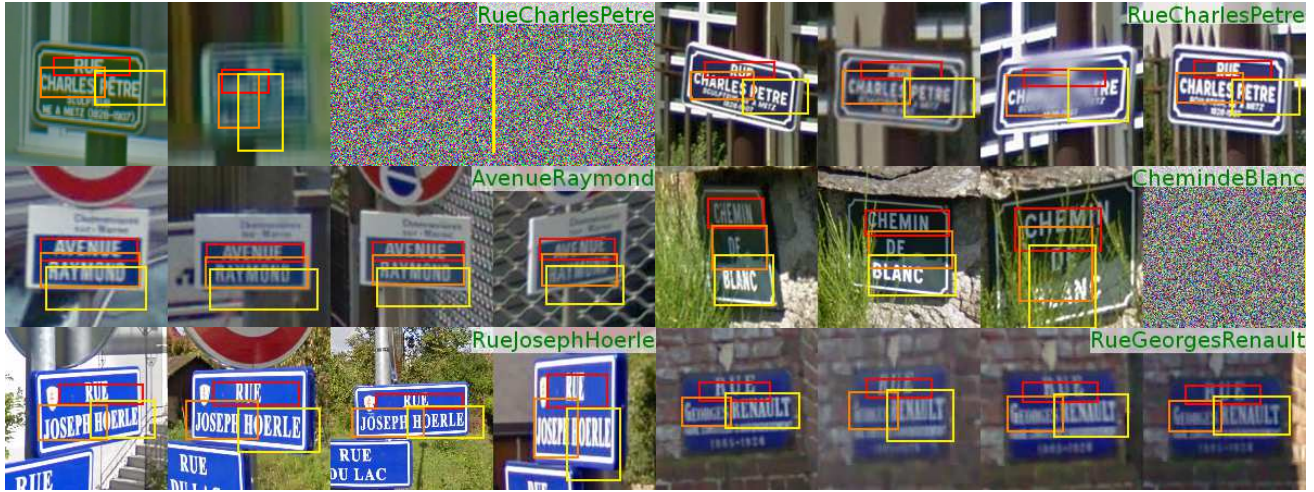}
	\end{center}
	\caption{Samples from the \ac{FSNS} dataset, these examples show that our system is able to detect a range of differently arranged text lines and also recognize the content of these words}
	\label{fig:fsns_examples}
\end{figure*}

\subsection{Preliminary Experiments on the \ac{FSNS} dataset}
\label{ssec:fsns_experiments}

Following our scheme of increasing the difficulty of the task that should be solved by the network, we chose the \acf{FSNS} dataset by Smith \etal \cite{Smith2016EndToEnd} to be our third dataset to perform experiments on.
The results we report here are preliminary and are only meant to show that our network architecture is also applicable to this kind of data, although it does not yet reach state-of-the-art results.
The \ac{FSNS} dataset contains images of french street name signs that have been extracted from Google Streetview.
This dataset is the most challenging dataset for our approach as it 
\begin{enumerate*}[label={(\arabic*)}]
	\item contains multiple lines of text with varying length embedded in natural scenes with distracting backgrounds and
	\item contains a lot of images that do not include the full name of the streets.
\end{enumerate*}

During our first experiments with that dataset we found that our model is not able to converge, when trained on the supplied groundtruth. We argue that this is because the labels of the original dataset do not include any hint on which words can be found in which text line.
We therefore changed our approach and started with experiments where we tried to find individual words instead of textlines with more than one word.
We adapted the groundtruth accordingly and used all images that contain a maximum of three words for our experiments, which leaves us with approximately \SI{80}{\percent} of the original data from the dataset.
\autoref{fig:fsns_examples} shows some examples from the \ac{FSNS} dataset where our model correctly localized the individual words and also correctly recognized the words.
Using this approach we were able to achieve a reasonably good character recognition accuracy of \SI{97}{\percent} on the test set, but only a word accuracy of \SI{71.8}{\percent}.
The discrepancy in character recognition rate and word recognition rate is caused by the fact that the model we trained for this task uses independent softmax classifiers for each character in a word.
Having a character recognition accuracy of \SI{97}{\percent} means that there is a high probability that at least one classifier makes a mistake and thus increases the sequence error.

	\section{Conclusion}
\label{sec:conclusion}

In this paper we presented a system that can be seen as a step towards solving end-to-end scene text recognition, using only a single multi-task deep neural network.
We trained the text detection component of our model in a semi-supervised way and are able to extract the localization results of the text detection component.
The network architecture of our system is simple, but it is not easy to train this system, as a successful training requires extensive pre-training on easier sub-tasks before the model can converge on the real task.
We also showed that the same network architecture can be used to reach competitive or state-of-the-art results on a range of different public benchmark datasets for scene text detection/recognition.

At the current state we note that our models are not fully capable of detecting text in arbitrary locations in the image, as we saw during our experiments with the \ac{FSNS} dataset.
Right now our model is also constrained to a fixed number of maximum textlines/characters that can be detected at once, in our future work we want to redesign the network in a way that makes it possible for the network to determine the number of textlines in an image by itself.


	{\small
		\bibliographystyle{ieee}
		\bibliography{paper}
	}
\end{document}